\documentclass[conference]{IEEEtran}
\usepackage[latin9]{inputenc}
\usepackage{amsmath}
\usepackage{graphicx}

\usepackage{times}
\usepackage{epsfig}
\usepackage{amsmath}
\usepackage{amssymb}
\usepackage{subfigure}
\usepackage{booktabs}
\usepackage{url}
\usepackage[bookmarks=false]{hyperref}
\usepackage{tabularx}
\usepackage{color}

\ifCLASSINFOpdf
\else
\fi

\usepackage{stfloats}
\begin{document}
%
\title{Generative Autoregressive Ensembles for Satellite Imagery Manipulation Detection}



\author{Daniel Mas Montserrat, J{\'a}nos Horv{\'a}th, S. K. Yarlagadda, Fengqing Zhu, Edward J. Delp\\ \\
Video and Image Processing Laboratory (VIPER)\\
School of Electrical Engineering\\
Purdue University\\
West Lafayette, Indiana, USA\\
}


%


\maketitle

\begin{figure}[b]
\vspace{-0.3cm}
\parbox{\hsize}{\em
WIFS`2020, December, 6-9, 2020, New York, USA.
978-1-7281-9930-6/20/\$31.00 \ \copyright 2020 IEEE.
}\end{figure}

\begin{abstract}
Satellite imagery is becoming increasingly accessible due to the growing number of orbiting commercial satellites. 
Many applications make use of such images: agricultural management, meteorological prediction, damage assessment from natural disasters, or cartography are some of the examples.
Unfortunately, these images can be easily tampered and modified with image manipulation tools damaging downstream applications.
Because the nature of the manipulation applied to the image is typically unknown, unsupervised methods that don't require prior knowledge of the tampering techniques used are preferred.
In this paper, we use ensembles of generative autoregressive models to model the distribution of the pixels of the image in order to detect potential manipulations.
We evaluate the performance of the presented approach obtaining accurate localization results compared to previously presented approaches.
\end{abstract}


%
\IEEEpeerreviewmaketitle

\section{Introduction}

Satellite imagery is used in a wide range of applications such as regional infrastructure levels assessment \cite{suraj_2017,oshri_2018}, agricultural crops classification \cite{russwurm_2019,brandt_2019}, forest characterization \cite{chauve_2009}, scene classification \cite{amirabbas_2017,shimoni_2008}, soil moisture estimation \cite{efremova_2018,alexakis_2017} and meteorological analysis, including precipitation prediction \cite{lebedev_2019}, thunderstorm detection \cite{zhang_2016_storm} and wind speed and direction estimation \cite{sahoo_2019}. These applications are possible thanks to the exponentially growing number of  commercial satellites \cite{ucs} (with many of those having imaging capabilities). 
Many image datasets captured by satellites are available to the public \cite{xia_2019,azimi_2019,yi_2010}, such as Planet Labs or the European Space Agency image datasets \cite{gupta_2019,schmitt_2019}.

Editing tools like GIMP \cite{gimp} or Photoshop \cite{photoshop_2016} can be used to forge and manipulate satellite images in a realistic manner. Furthermore, manipulation generation can be automated by using machine learning techniques \cite{nam_2018}, removing the need for manual editing. Such manipulation methods, combined with the ease of sharing data on the internet, can difficult the institutions and companies that make use of images captured by satellites. Indeed, several instances of manipulated images have surged in recent years, including the nighttime flyovers of India during the Diwali festivals \cite{byrd_2018}, the Malaysia Airlines Flight incident \cite{kramer_2016}, and the images of the spliced fake Chinese bridge \cite{edwards_2019}.

There is a wide range of manipulations techniques that can be used to forge satellite images. Some examples include splicing~\cite{cozzolino_2020} (cropping and pasting regions from different image sources), copy-move~\cite{barni2019copy} (cropping and pasting regions within the same image), shadow removal~\cite{yarlagadda2019shadow}, and machine learning-based forgeries, often generated using Generative Adversarial Networks (GANs) \cite{nam_2018}. 
Multiple methods to detect image manipulations have been proposed in recent years \cite{anderson_2011,schetinger_2017, bartolini_2001}. However, these methods are typically designed for images captured with consumer cameras and fail with images from other imaging devices, such as satellite imagery, with different compression schemes, post-processing, sensors, and color channels. 
Therefore, the detection of manipulations within satellite imagery still remains an unsolved problem that requires the development of new detection techniques that are accurate regardless of the nature of the manipulations and image capturing technology.

\begin{figure*}[!htpb]
\centering
\includegraphics[width=0.9\textwidth]{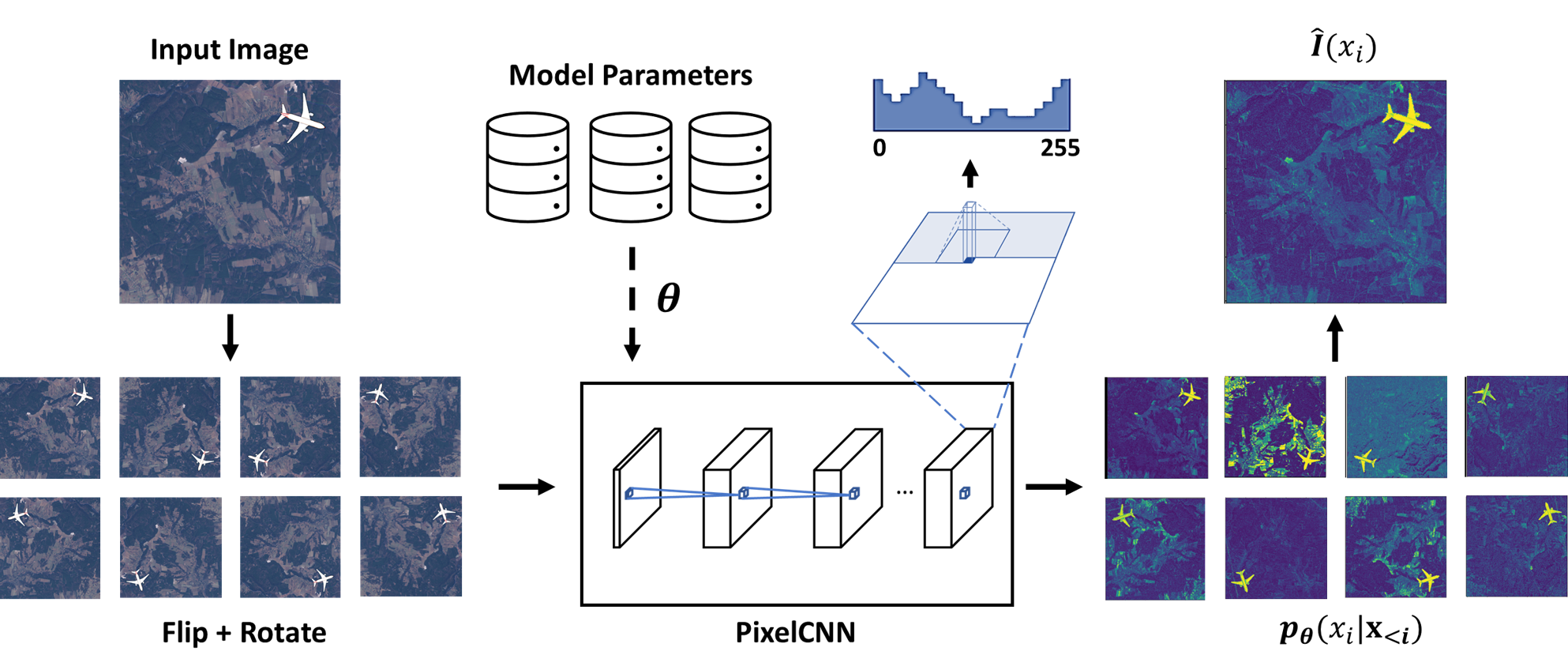}
\caption{Proposed ensemble: multiple models process the flipped and rotated input images. The prediction of every network is averaged obtaining a final robust and accurate likelihood estimate for each pixel of the image. The 8 images of $p_{\theta}(x_i|\mathbf{x}_{<i})$ are plotted as $-\log p_{\theta}(x_i|\mathbf{x}_{<i})$ for visualization purposes.}
\label{fig:method}
\end{figure*}

In this work, we show how PixelCNN \cite{oord2016pixel} and Gated PixelCNN \cite{van2016conditional}, two generative autoregressive models, can be used to detect pixel-level manipulations. These neural networks, commonly used to generate new images, can model the distribution of a pixel given a set of previously seen pixels (neighboring pixels). These neural networks can assign a conditional likelihood value to a given pixel, and in turn, a likelihood value to a complete image. Through sampling from the pixel distribution, new images can be generated in a sequential fashion. Furthermore, manipulated pixels can be detected by selecting the pixels with a low likelihood assigned by the neural network. By averaging the likelihood estimated by an ensemble of multiple networks, the method is able to obtain a more accurate manipulation localization. Figure \ref{fig:method} presents the proposed ensemble where multiple networks process the input image and its flipped and rotated versions. Then, all predictions are averaged in order to obtain a robust prediction. Finally, we evaluate the localization precision of the presented method using a dataset composed of images with splicing forgeries, first introduced in \cite{horvath_2020}.

The paper is organized as follows. In section \ref{sec:related-work} we present previous work on manipulation detection and autoregressive models. In section \ref{sec:dataset} we describe the dataset composed by images captured by a satellite. In section \ref{sec:method} we describe the presented method. In section \ref{sec:experimental-results} we show the experimental results and we conclude the paper with section \ref{sec:conclusions}.

\section{Related Work} \label{sec:related-work}

Many techniques to detect a wide range of image manipulations have been previously presented.
Some examples include techniques that detect manipulations by using embedded meta-data \cite{Guera2019_ICMLW}, finding double-JPEG compression artifacts \cite{barni_2010}, using neural networks with domain adaptation \cite{cozzolino_2018}, using deepfakes detection neural networks \cite{montserrat_2020} or using saturation cues \cite{mccloskey_2019}.
Several methods have proved to accurately detect spliced objects within images captured with consumer-level cameras \cite{cozzolino_2020, cozzolino_2015}.
The method presented in \cite{cozzolino_2020} extracts a fingerprint from the camera model used to capture the image in order to suppress the scene content and enhance camera model-related artifacts.
The method presented in \cite{cozzolino_2015} makes use of a feature-based technique that can detect splicing in images without any prior information of the nature of the manipulations by detecting traces left locally by processing steps within the capturing device.
Unfortunately, many of these methods perform poorly when applied to satellite imagery. The image acquisition process differs between consumer cameras (including smartphone cameras) and satellites: different sensor technologies and post-processing steps such as orthorectification, radiometric corrections, and compression are used. Because of these differences, methods designed for consumer cameras do not transfer properly to satellite imagery. 

Recently, multiple methods designed to detect forgeries in satellite imagery have been introduced. These include methods using hand-crafted features, like watermarking-based techniques \cite{ho_2005}, and data-driven machine learning-based approaches including supervised \cite{bartusiak_2019} and unsupervised \cite{sri_2018, horvath_2019, horvath_2020} methods. 
While supervised methods tend to perform better, they might not generalize well to types of manipulations that were not present in the training set. Therefore, unsupervised methods, which don't make use of manipulated data during training, are preferred.
The supervised method presented in \cite{bartusiak_2019} makes use of a conditional GAN 
to detect and localize splicing forgeries in satellite images by estimating a forgery mask.
The work introduced in \cite{sri_2018} is based on a GAN that encodes patches from the input image into a low dimensional vector that is later used by a one-class support vector machine (SVM) to detect if a patch contains forgeries or not.
The method presented in \cite{horvath_2019}, named Sat-SVDD, is a kernel-based one-class classification method that detects splicing forgeries by using a modified Support Vector Data Description (SVDD) \cite{tax_2004}.
The SVDD encodes each patch from the original images (without manipulations) to a latent space within a hypersphere. During testing, the latent vectors that are placed outside the hypersphere are considered as patches containing a forgery.
The method in \cite{horvath_2020} makes use of a deep belief network (DBN)  \cite{hinton_2006_b} composed of two stacked layers of restricted Boltzmann machines (RBM) \cite{smolensky_1986} parametrized with uniform distributions. The deep belief network is used to reconstruct patches extracted from the image. Then, the reconstruction error is used to detect if manipulations are present: patches with a reconstruction error higher than a threshold are considered as forgeries.

 In this work, we use generative autoregressive models, specifically PixelCNN \cite{oord2016pixel} and Gated PixelCNN \cite{van2016conditional}, which are described in the following sections. Many autoregressive generative models have been presented in recent years \cite{germain2015made, 
 van2016conditional, oord2016pixel}. Autoregressive models are able to estimate the distribution of an image by estimating the conditional distribution of each pixel. The distribution of each pixel is estimated given its neighboring pixels. Then, the distribution of an image can be expressed as the product of the conditional distributions. These models make use of masked convolutions in order to respect autoregressive constraints: each pixel is reconstructed only from previous pixels in a given ordering. PixelCNN and its recurrent-based counterpart PixelRNN \cite{oord2016pixel}, showed that autoregressive modeling can be successfully used to generate new images. Many variations have been presented such as Gated PixelCNN \cite{van2016conditional}, PixelCNN++ \cite{salimans2017pixelcnn++}, PixelSNAIL \cite{chen2017pixelsnail}. Furthermore, the same approach has been extended to video modeling in Video Pixel Network (VPN) \cite{kalchbrenner2017video}, variational autoencoders in PixelVAE \cite{gulrajani2016pixelvae} and PixelVAE++ \cite{sadeghi2019pixelvae++}, and to generative adversarial networks in PixelGAN \cite{makhzani2017pixelgan}.

Some works have studied the capability of likelihood models to detect outliers. The work presented in \cite{choi2018waic} makes use of the Watanabe-Akaike Information Criterion (WAIC) 
to detect outliers. The work in \cite{serra2019input} normalizes the likelihood estimate of an image with a measure of complexity to detect outliers. 
However, most of these approaches focus on image-level out of distribution (OoD) estimates (also referred to as anomaly detection), and likelihood methods to detect pixel-level manipulations remain unexplored.

\section{Dataset} \label{sec:dataset}

\begin{figure}[!thpb]
\centering
\includegraphics[width=0.9\columnwidth]{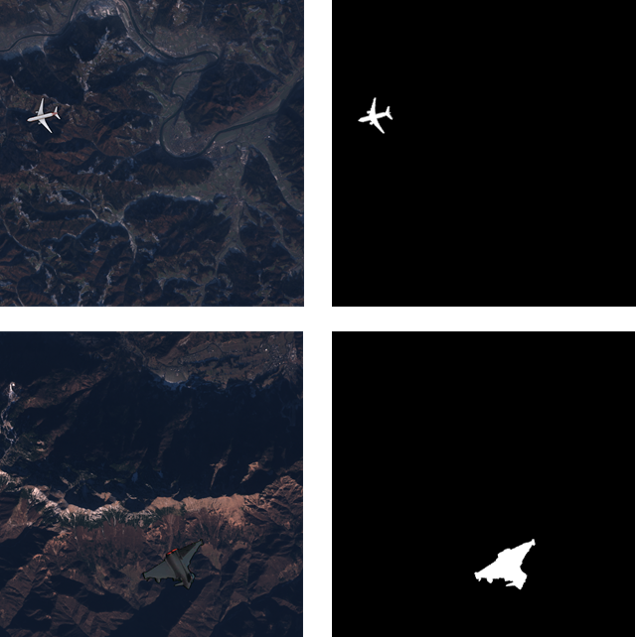}
\caption{Examples of images (left) from the dataset and its corresponding manipulation masks (right).}
\label{fig:dataset}
\end{figure}

In order to train and evaluate our method, we use the dataset first introduced in \cite{horvath_2020}. This dataset is composed of orthorectified satellite images including regions of Slovenia taken from the Sentinel program \cite{Sentinel}. The images have a resolution of $1000\times1000$ pixel. We use a subset of the dataset consisting of 98 original images (without manipulations) for training and 500 manipulated images with their corresponding ground truth masks for testing. Each manipulated image has one spliced object randomly selected among 19 different objects, including clouds, planes, smoke, and drones. The objects are spliced with different locations, rotation angles and sizes including $16\times16$, $32\times32$, $64\times64$, $128\times128$, and $256\times256$ pixels. Figure \ref{fig:dataset} presents some examples of the dataset.

\section{Proposed Method} \label{sec:method}

\subsection{Autoregressive Models}
PixelCNN \cite{oord2016pixel} and Gated PixelCNN \cite{van2016conditional} are neural networks composed by multiple fully-convolutional residual layers and are trained to model the distribution of an image $\mathbf{x}$ as the product of the conditional distributions of every pixel $x_{i}$:


\begin{equation}
\label{eq:1}
p(\mathbf{x}) = \displaystyle\prod_{i=1}^{L} p(x_i|x_1,...,x_{i-1})
\end{equation}

Where $\mathbf{x}$ is an image of $L$ pixels and $x_{i}$ is the $i$th pixel of the image. The predicted distribution of every pixel $x_{i}$ is conditional to the previous pixels $x_1,...,x_{i-1}$ in a raster scan order: row by row and pixel by pixel within every row (left to right and top to bottom). 


In RGB images, each color channel (R, G, B) is modeled successively: first the red channel, then the green channel conditioned to the red, and finally the blue channel conditioned to the red and green. Therefore, the conditional probability of an RGB pixel is as follows:

\begin{equation}
\label{eq:2}
\begin{split}
p(x_i|\mathbf{x}_{<i}) = p(x_i|x_1,...,x_{i-1}) = \\ p(x_{i,R}|\mathbf{x}_{<i})p(x_{i,G}|\mathbf{x}_{<i},x_{i,R})p(x_{i,B}|\mathbf{x}_{<i},x_{i,R},x_{i,G})
\end{split}
\end{equation}

The autoregressive constraints are achieved by masking the convolutions accordingly, both within spatial dimensions and within features maps. The use of convolutions allows the network to perform the likelihood predictions in parallel during training and testing but the image generation remains a sequential process.

While our method is designed for RGB images, it is common for satellites to capture multi-spectral images containing more than 3 channels. The presented approach can be easily extended to any number of channels by assigning some arbitrary order within the channels and estimating the conditional probability as follows:

\begin{equation}
\label{eq:3}
p(x_i|\mathbf{x}_{<i}) = \displaystyle\prod_{j=1}^{C} p(x_{i,j}|\mathbf{x}_{<i},x_{i,1},...,x_{i,{j-1}})
\end{equation}

Where $x_{i,j}$ is the $i$th pixel from the $j$th channel of an image with a total of $C$ channels.

PixelCNN and Gated PixelCNN models the conditional probability $p(x_i|\mathbf{x}_{<i})$ as a multinomial (categorical) distribution through a softmax layer where each channel within the image can take a value from 0 to 255.
The network takes as input an image with $N \times M \times 3$ dimensions (with $N \times M = L$) and outputs a prediction with dimension $N \times M \times 3 \times 256$. 
While the original PixelCNN is designed to work with 8-bit images, the method can be adapted to work with images with different bit depths by properly changing the range of values that the softmax layer can take. For example, when working with 11-bit images, the softmax layer should output values from 0 to 2047. This is especially useful for satellite imagery as many datasets have bit depths higher than 8-bits.

\subsection{Generative Ensembles}

\begin{figure}[!t]
\centering
\includegraphics[width=0.9\columnwidth]{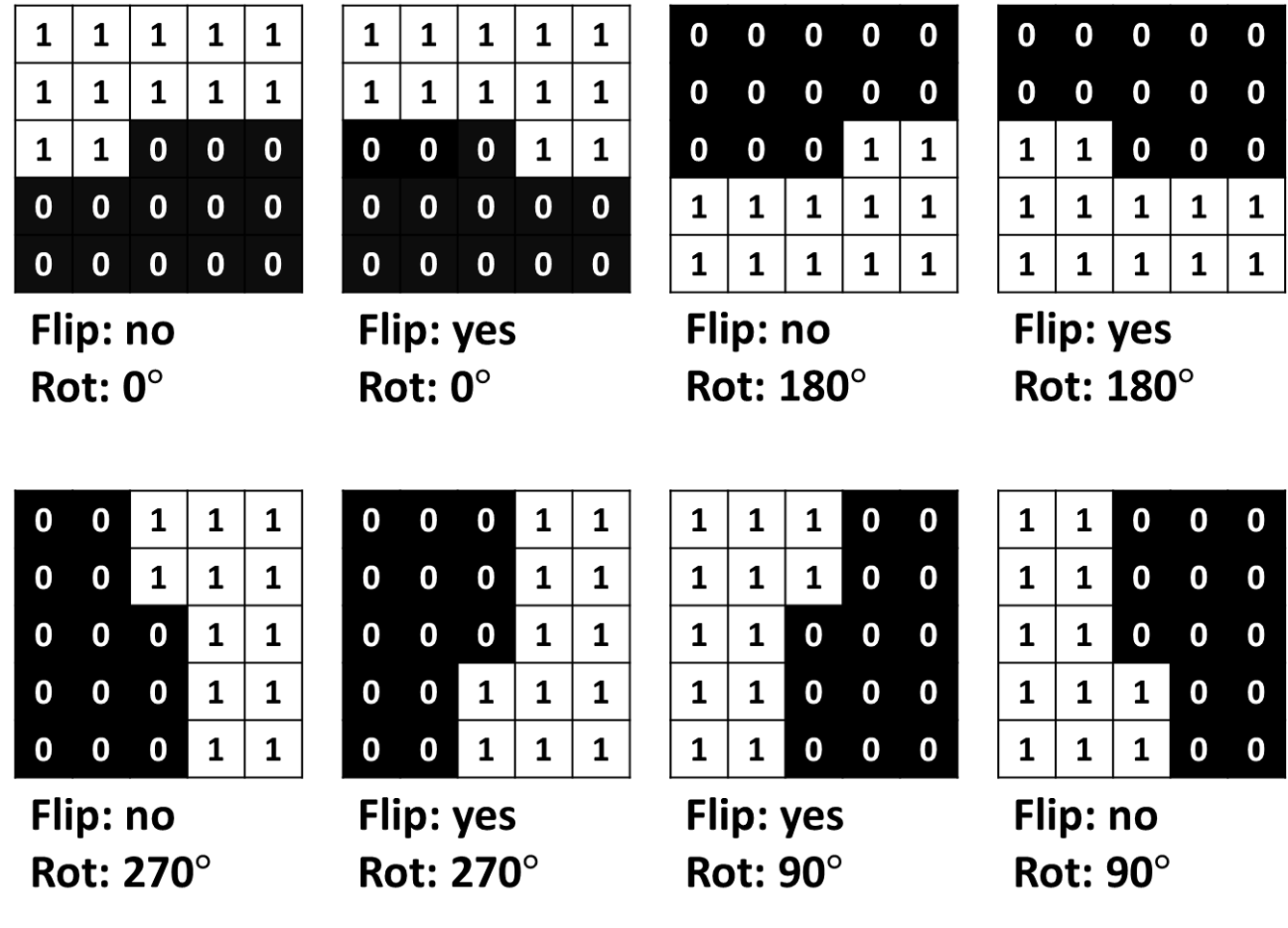}
\caption{Different convolutional masks and the respective transformation (flipping and rotation) performed to the input image to obtain the equivalent effect.}
\label{fig:masks}
\end{figure}

We can obtain more accurate and robust predictions by combining multiple networks within an ensemble. We average the predictions of multiple networks with different parameters and scan orderings. In order to obtain multiple model parameters, we save the parameters of the network at different epochs during the training process. 
The parameters $\theta$ of the network at each training epoch can be seen as an approximate proxy of posterior samples of $p(\theta|\mathcal{D})$ (the distribution of the model parameters given the training set $\mathcal{D}$). 
To use different scan orderings during the autoregressive modeling (the order in which neighboring pixels are observed) we can apply different masks to the convolutional filters, or equivalently, rotate and horizontally flip the input image. Figure \ref{fig:masks} shows the 8 different orderings used and the transformations (flip and rotate) applied to the input image and the corresponding convolutional mask to obtain equivalent results.

The average of the prediction of multiple networks with different parameters and scan order can be understood as a Monte Carlo approximation of the marginal likelihood of each pixel $p(x_{i})$, where the effect of the model parameters $\theta$ and scan ordering $\mathbf{x}_{<i}$ are smoothed out:



\begin{equation}
\label{eq:4}
p(x_i) = \mathbb{E}_{\theta,\mathbf{x}_{<i}}[ p_{\theta}(x_i|\mathbf{x}_{<i})] \approx \frac{1}{K} \sum_{\omega \in \Omega} p_{\theta}(x_i|\mathbf{x}_{<i})
\end{equation}

Where $\omega$ are samples of model parameters and scan ordering pairs 
$(\theta,\mathbf{x}_{<i})$ from a set $\Omega$ of size $|\Omega|=K$.
In other words, the average of the prediction of $K$ networks with different parameters and scan orderings are used to approximate the marginal likelihood $\hat{p}(x_i) \approx p(x_i)$. 
In order to detect manipulations, we can use the negative log-likelihood, which in turn is the information content (or Shannon information) quantity $I(x_i) = -\log p(x_i)$, approximated as:



\begin{equation}
\label{eq:4}
\hat{I}(x_i) =  -\log [ \frac{1}{K} \sum_{\omega \in \Omega} p_{\theta}(x_i|\mathbf{x}_{<i})]
\end{equation}


A pixel is considered to be manipulated if $\Hat{I}(x_{i}) > T$, where $T$ is experimentally selected.
Ideally, the model will assign high likelihood values (and thus small information values) to pixels that have not been manipulated, and small likelihood (and high information) values to manipulated pixels.


\subsection{Training and Testing Setup}
In this work, we use a PixelCNN composed of 7 residual blocks and a Gated PixelCNN composed of 6 gated blocks. We train the networks with the Adam optimizer with a learning rate of 0.001. During the training process, we randomly rotate 0, 90, 180, or 270 degrees and horizontally flip the images. We train the network for 1000 epochs and we store the model parameters every 20 epochs.

During testing, the ensemble is composed of $K=50$ different models. We select 50 model parameters uniformly distributed from epoch 30 to epoch 1000 of the training process. For each model parameters we use a scan ordering randomly selected from the 8 different scan orderings shown in figure \ref{fig:masks}.

\section{Experimental Results} \label{sec:experimental-results}

We train and evaluate our method with the dataset presented in Section \ref{sec:dataset}. 
The ensemble of networks is trained only with original images and no manipulated images are used during the training process. 
In order to evaluate the localization performance of the presented method, we compute the area under the curve (AUC) of the Precision/Recall (P/R) curves by changing the threshold $T$ applied to the estimated negative loglikelihood $\Hat{I}(x_{i})$. 
Table \ref{table:detection_satellite} presents our results compared with previous methods. 
Different Precision/Recall are shown for each of the different sizes of the objects spliced into the images. 
For example, P/R$_{32}$ is the AUC of the P/R curve for manipulated images with spliced objects of size $32\times32$. 

Our experimental results show that the generative ensemble of PixelCNNs and Gated PixelCNNs outperform previously presented methods. 
Methods \cite{cozzolino_2020} and \cite{cozzolino_2015} have been previously used to successfully detect splicing manipulations in consumer-level cameras but our experiments suggest that they fail when faced with satellite imagery.
While most of the methods fail to detect objects smaller than $64\times64$, the presented generative ensembles are able to properly detect small forgeries.
Methods such as \cite{sri_2018}, \cite{horvath_2019}, and \cite{horvath_2020} produce estimates within patches of the input image, therefore lacking enough resolution to detect small forgeries. On the contrary, PixelCNN and Gated PixelCNN process the whole image in a fully-convolutional manner and detects manipulations in a pixel-level providing higher detection accuracy.
We can observe that Gated PixelCNN provides more accurate results than the regular PixelCNN network, especially for objects smaller than $64 \times 64$ pixels. These results are aligned with previous works \cite{van2016conditional} which have shown that Gated PixelCNN is able to model the image distribution of the training images more accurately (with a lower negative log-likelihood score) than PixelCNN.

Note that the networks are trained in a fully unsupervised manner without any semantic information (label or ground truth) of the spliced objectes. Therefore, the networks learn the regularities of the training images and detect as manipulations out-of-distribution regions of the images, regardless of the spliced object (e.g. clouds, planes...). While this approach is well suited for overhead imagery where large (unsupervised) datasets can be collected and images are visually similar (e.g. mountain regions tend to look alike from overhead imagery), further experiments need to be performed on consumer-level camera multimedia images.

\begin{table*}[!htpb]
	\centering
	\caption {AUC scores (\%) of the P/R curves for the localization task. The subscript (P/R$_{\times}$) denotes the manipulation size.}
	\label{table:detection_satellite}
	\begin{tabular}{lcccccc}

	Method & P/R$_{16}$  &	P/R$_{32}$  & P/R$_{64}$  & P/R$_{128}$ & P/R$_{256}$ & Average \\
	\toprule
	Noiseprint~\cite{cozzolino_2020} & $0.0$ & $0.1$ & $2.5$ & $4.6$ & $7.8$ & $3.0$ \\
	Yarlagadda \textit{et al}~\cite{sri_2018} & $0.0$ & $0.3$     & $2.5$     & $18.3$     & $37.8$  & $11.7$    \\
	Splicebuster~\cite{cozzolino_2015} &  $0.0$ &  $0.5$     & $7.8$     & $31.2$     & $48.5$  & $17.6$  \\
	Sat-SVDD~\cite{horvath_2019} & $0.1$ & $1.4$ & $18.1$ &  $34.4$ & $55.7$  & $21.9$ \\
	UU-DBN~\cite{horvath_2020} & $7.5$ & $13.3$ & $31.7$ & $40.5$ & $48.8$ & $28.4$\\
	\midrule
	Generative Ensemble (PixelCNN) & $37.6$ & $44.6$ & $56.2$ & $65.3$ & $\mathbf{75.6}$ & $55.9$ \\
	Generative Ensemble (Gated PixelCNN) & $\mathbf{46.3}$ & $\mathbf{53.8}$ & $\mathbf{61.1}$ & $\mathbf{65.6}$ & $72.8$ & $\mathbf{59.9}$\\
	\bottomrule
	\end{tabular}

\end{table*}

\section{Conclusions} \label{sec:conclusions}
The wide range of manipulations that can be applied to images and the large diversity of imaging technologies used in satellites makes their detection a challenging problem that still remains unsolved.
In this paper, we introduced an unsupervised splicing detection method. The method consists of an ensemble of generative autoregressive models that estimates the pixel distribution of the image. The method is capable to accurately detect manipulated pixels by selecting the regions of the image where the network predicts a low likelihood value.
The presented method is fully unsupervised and doesn't use any prior knowledge from the applied manipulation during training. Our experiments show that the localization accuracy of our method surpasses the previous works and shows that generative models, specially autoregressive-based networks, provide a promising approach to detect pixel-level manipulations.

\section{Acknowledgment}
This material is based on research sponsored by DARPA and Air Force Research Laboratory (AFRL) under agreement number FA8750-16-2-0173. 
The U.S. Government is authorized to reproduce and distribute reprints for Governmental purposes notwithstanding any copyright notation thereon. 
The views and conclusions contained herein are those of the authors and should not be interpreted as necessarily representing the official policies or endorsements, either expressed or implied, of DARPA and Air Force Research Laboratory (AFRL) or the U.S. Government.

Address all correspondence to Edward J. Delp, ace@ecn.purdue.edu .

\bibliographystyle{IEEEbib}

\bibliography{egbib}
%

%
%
%
%
%

\end{document}